\documentclass{article}

\usepackage{arxiv}

\usepackage[utf8]{inputenc} 
\usepackage[T1]{fontenc}    
\usepackage{hyperref}       
\usepackage{url}            
\usepackage{booktabs}       
\usepackage{amsfonts}       
\usepackage{nicefrac}       
\usepackage{microtype}      

\usepackage{lipsum}
\usepackage{multirow}
\usepackage{subcaption}
\usepackage{hhline}
\usepackage{graphicx}

\usepackage{xcolor}
\definecolor{RoyalBlue}{RGB}{65, 105, 225}

\title{Exploring the Potential of Large Language Models
in Public Transportation: San Antonio Case Study}

\author{
    Ramya Jonnala\hspace{3mm} 
    Gongbo Liang\hspace{3mm} 
    Jeong Yang\hspace{3mm}
    Izzat Alsmadi \\ [0.75ex]
    Department of Computational, Engineering, and Mathematical Sciences \\
    Texas A\&M University-San Antonio\\ [0.75ex]
   \texttt{ rjonn01@jaguar.tamu.edu,~\{gliang, jyang, ialsmadi\}@tamusa.edu}
}
		
\begin{document}
\maketitle

\begin{abstract}
    The integration of large language models (LLMs) into public transit systems presents a transformative opportunity to enhance urban mobility. This study explores the potential of LLMs to revolutionize public transportation management within the context of San Antonio's transit system. Leveraging the capabilities of LLMs in natural language processing and data analysis, we investigate their capabilities to optimize route planning, reduce wait times, and provide personalized travel assistance. By utilizing the General Transit Feed Specification (GTFS) and other relevant data, this research aims to demonstrate how LLMs can potentially improve resource allocation, elevate passenger satisfaction, and inform data-driven decision-making in transit operations. A comparative analysis of different ChatGPT models was conducted to assess their ability to understand transportation information, retrieve relevant data, and provide comprehensive responses. Findings from this study suggest that while LLMs hold immense promise for public transit, careful engineering and fine-tuning are essential to realize their full potential. San Antonio serves as a case study to inform the development of LLM-powered transit systems in other urban environments. 
    \let\thefootnote\relax\footnotetext{This work is accepted to AAAI 2025 Workshop on AI for Urban Planning.}
\end{abstract}

\keywords{LLMs \and Neural Network \and Pre-Training \and Urban Planning \and Artificial Intelligence}

\section{Introduction}

The rise of artificial intelligence (AI) and machine learning (ML) has initiated a new era of technological advancements, revolutionizing numerous sectors, such as cyber security~\cite{alsmadi2022adversarial,ahmad2022deep,liang2023enhancing}, healthcare~\cite{liang2019joint,xing2023self,liu2023simulated}, and public transportation~\cite{chen2021safety,alsrehin2023u2,liang2023unveiling}. Among these innovations, large language models (LLMs), such as OpenAI's GPT series~\cite{chatgpt2022,achiam2023gpt}, have demonstrated exceptional natural language processing, understanding, and generation capabilities. These models can analyze vast amounts of data~\cite{kalla2023study,alyasiri2023exploring}, generate human-like text~\cite{orru2023human}, and facilitate complex decision-making processes~\cite{goossens2023integrating,li2023revolutionizing}, making them potentially invaluable tools for enhancing public transit systems. 

Public transit systems are the backbone of urban mobility, delivering essential services to millions of passengers every day~\cite{abdallah2023sustainable,xu2022urban}. Efficient and reliable public transportation is crucial for reducing traffic congestion, minimizing environmental impact, and promoting equitable access to mobility~\cite{guo2020systematic}. However, transit agencies frequently encounter challenges such as fluctuating passenger demand, optimizing routes, maintaining real-time communication with passengers, and efficiently allocating resources~\cite{zhang2021agent,huang2020flexible}. Traditional methods of addressing these issues may need to be revised due to their limited scalability and adaptability.

San Antonio, one of the fastest-growing cities in the United States, offers a unique case study for exploring the integration of LLMs in public transit. The city's rapid population growth has heightened the need for efficient public transportation solutions~\cite{altamirano2024to,houston2023san}. As the local transit authority seeks innovative methods to enhance service delivery, deploying LLMs presents a promising solution for addressing current and future challenges.

This study aims to explore the potential of LLMs to enhance various aspects of San Antonio's public transit system. The following are key areas where LLMs could be beneficial in public transportation:
\begin{itemize}
    \item Optimize Route Planning and Scheduling: Analyze historical and real-time data to enhance route planning and scheduling, thus reducing wait times and improving service reliability.
    \item Enhance Passenger Communication: Use of LLMs for real-time engagement with passengers, offering personalized travel assistance, updates, and recommendations.
    \item Improve Operational Efficiency: LLMs can influence resource allocation, including the deployment of buses and drivers, to boost overall operational efficiency.
\end{itemize}

\section{Significance of the Study}
\label{sec:headings}
The integration of LLMs into public transit systems has the potential to transform urban mobility by enhancing efficiency, responsiveness, and user-friendliness. This study not only advances academic understanding of AI applications in transportation but also offers practical insights for transit authorities and policymakers. By examining San Antonio—a city representative of many growing urban areas—the findings can be applied to other cities facing similar challenges.

Additionally, the research underscores the broader implications of AI in public services, highlighting the significance of ethical considerations, data privacy, and the necessity for ongoing evaluation and improvement. As cities around the world navigate the complexities of modern urbanization, the lessons learned from San Antonio's experience with LLMs can provide valuable guidance for future innovations in public transit systems.

In summary, this study seeks to bridge the gap between cutting-edge AI technologies and their practical applications in public transportation, illustrating how LLMs can be utilized to develop smarter, more adaptive, and passenger-focused transit networks. The following sections will explore the theoretical framework, detailed methodology, findings, and implications of this transformative approach to public transit management.

\section{Related Work}
The integration of LLMs such as OpenAI's GPT-4~\cite{achiam2023gpt} into public transit systems is a burgeoning field that aims to enhance the efficiency, accessibility, and user experience of public transportation~\cite{alyasiri2023exploring}. LLMs can process and analyze vast amounts of data~\cite{li2023revolutionizing}, generate human-like text~\cite{orru2023human}, and understand complex queries~\cite{li2023revolutionizing}, making them suitable for a range of applications in public transit. This literature review explores the current state of research on the deployment of LLMs in public transit systems, focusing on areas such as passenger information services, operational efficiency, and accessibility improvements.

One of the primary applications of LLMs in public transit is in improving passenger information services. Studies have demonstrated that LLMs can enhance the quality and accuracy of real-time information provided to passengers. For instance, researchers explored the use of GPT in generating real-time updates and personalized travel advice for passengers~\cite{papangelis2020plato,yenduri2024gpt,voss2023bus,khalil2024advanced}. Their findings indicated that LLMs could effectively handle complex passenger queries and provide accurate, context-aware responses, thereby improving the overall passenger experience.

Furthermore, researchers highlighted the potential of LLMs in multilingual support for transit systems~\cite{ullah2024role,kaur2024text,zheng2023trafficsafetygpt}. Given the diverse linguistic backgrounds of urban populations, LLMs like GPT-4 can be trained to provide information in multiple languages, ensuring that non-native speakers have equal access to transit information. This capability not only improves user satisfaction but also promotes inclusivity and accessibility.

The paper,~\cite{devunurichatgpt}, presents an evaluation of large language models (LLMs), specifically ChatGPT, in interpreting and retrieving information from General Transit Feed Specification (GTFS) data. The study demonstrates that ChatGPT can effectively understand and respond to various queries about public transit schedules and services, showcasing its potential in enhancing transit information systems. However, the paper also highlights areas for improvement, such as the model's occasional inaccuracies and the need for further fine-tuning to handle complex and domain-specific transit queries more reliably.

The paper,~\cite{zheng2023chatgpt}, explores the potential of using ChatGPT and similar large language models (LLMs) to revolutionize intelligent transportation systems. It argues that LLMs could significantly enhance various aspects of transportation, such as traffic management, passenger assistance, and operational efficiency, but also points out the challenges related to data privacy, model accuracy, and integration with existing systems.

\section{Goals}
\label{sec:goal} 



Contemporary large language models predominantly employ learning-based approaches. Prominent examples include ChatGPT~\cite{chatgpt2022}, built upon the Transformer architecture~\cite{vaswani2017attention} and trained using generative pre-training techniques~\cite{radford2018improving,radford2019better,brown2020language}. The performance of these models is inherently contingent upon the quality and quantity of their training data. Consequently, erroneous LLM outputs can arise from various factors, including insufficient training data on specific topics or architectural limitations in processing user input, such as inadequate embedding methods. To effectively evaluate learning-based LLMs, it is imperative to distinguish between pre-trained models and underlying architectures.

This project aims to evaluate LLMs' capacity to comprehend GTFS and other public transportation information through two experimental approaches:
\begin{enumerate}
    \item \textbf{Performance of Pre-trained Models:} We will assess the ability of a pre-trained LLM ``as-is" to answer transportation-related questions without additional context. This evaluation will determine the model's inherent understanding of GTFS and public transit information. Errors in this phase may indicate limitations in the model's pre-training data or architectural constraints related to processing transportation-specific queries. We refer to this as the ``understanding" task.
    
    \item \textbf{Impact of LLM Architecture:} To isolate the impact of LLM architecture, we will provide LLMs with explicit GTFS data and public transportation information before posing transportation-related questions. By question-answer tests that involve finding answers from the provided GTFS data, we can determine whether the initial failures were due to information deficits or architectural limitations. We refer to this as the ``information retrieval" (IR) task.    
    
\end{enumerate}

The findings from these tasks will offer valuable insights into the cause of errors. For instance, if the LLMs can answer the questions correctly in the second experiment but not the first, it suggests insufficient pre-training data on the specific topic within the models. Conversely, the results might indicate that even with adequate data, the LLM models struggle with the questions, potentially due to architectural limitations.

\section{Approaches and Experiment Design}
In this project, we employ OpenAI's ChatGPT as the representative LLM due to its widespread public availability through both a web portal and a programmatic API. To assess LLM capabilities in understanding and retrieving transportation information (Goals 1 and 2, respectively, as outlined in Section~\ref{sec:goal}), we conducted five experiments comprising 3275 multiple-choice questions and 80 short-answer questions based on San Antonio's public transportation system.

\subsection{Experiments for Transportation Understanding}
The transportation information understanding task, referred to as \textit{understanding}, evaluates a pre-trained LLM's ability to comprehend and answer questions about San Antonio's public transportation system.  

Following~\cite{devunurichatgpt}, we designed 195 multiple-choice questions (MCQs) with a single correct answer,  meticulously crafted to cover six key question categories (Table~\ref{table:questions}). The questions are derived using the official GTFS Schedule documentation~\footnote{\href{https://gtfs.org/schedule/reference}{https://gtfs.org/schedule/reference}} and used in the initial experiment (\texttt{Experiment I}) to assess the LLM's understanding of transportation information. 

\begin{table}
    \centering
    \caption{GTFS Understanding Benchmarking dataset questionnaire and their categories}
    \small
    \begin{tabular}{c|c} \toprule \hline
         \textbf{Question Type} & \textbf{Number of Questions}\\ \hline\hline
         Term Definitions& 14\\ \hline
         Common Reasoning& 28\\ \hline
         File Structure& 17\\ \hline
         Attribute Mapping& 32\\ \hline
         Data Structure& 30\\ \hline
         Categorical Mapping& 74\\ \hline
         Total& 195\\ \hline
    \bottomrule
    \end{tabular}
    \label{table:questions}
\end{table}

\begin{table*}[!tb]
\centering
\caption{Ten sample MCQs that are used in this study}
\small
\begin{tabular}{  l |c || p{11cm} } \toprule \hline
  \textbf{Category} &  \textbf{Type} & \textbf{Question} \\ \hline\hline
    
    Categorical Mapping& Original & In the ``trips.txt" file, what is the meaning of ``wheelchair\_accessible" 0 or empty? a) No accessibility information for the trip b) Vehicle being used on this particular trip can accommodate at least one rider in a wheelchair c) No riders in wheelchairs can be accommodated on this trip d) Stop cannot be accessed by anyone A question \\ \hline
    Attribute\ Mapping & Original & In which file does the shape\_dist\_travelled attribute\ appear in GTFS? a) stops.txt b) shapes.txt c) trips.txt d) stop\_times.txt A question \\ \hline
    Common Reasoning& Original& Can a GTFS feed contain multiple agency information? a) Each agency should publish a separate GTFS. b) No, GTFS feeds can only represent a single agency. c) Multiple agency information is specified in the "agency.txt" file. d) Agencies are not relevant in GTFS feeds. A question \\ \hline
    Data Structure& Original& How is the wheelchair\_accessible attribute represented in GTFS? a) Boolean (true or false) b) Float (number of accessible seats) c) Enum (e.g., 0,1,2) d) Text representation of wheelchair accessibility ...\\ \hline
    File Structure& Original& What is the purpose of the ``transfers.txt" file in GTFS? a) It contains information about fare rules and transfers. b) It provides details about the geographic shapes of routes. c) It specifies the frequency of trips. d) It provides real-time arrival and departure information. \\\hline
    Term Definition& Original&What is a dataset in the context of GTFS? a) A single file containing all transit information b) A collection of tables representing different entities c) A specific date for transit service d) A record representing a transit agency \\ \hline
    Attribute\ Mapping& Augmented& In which file can you find the route\_desc attribute in GTFS? a) stops.txt b) None of these c) trips.txt d) calendar.txt \\ \hline
    Categorical Mapping& Augmented& What value is used in the ``wheelchair\_boarding" field of the "stops.txt" file to indicate that the stop has no information regarding wheelchair accessibility? a) 0 b) 1 c) None of these d) 3 \\ \hline
    Common Reasoning& Augmented& How does GTFS handle multiple trips on the same route at the same time? a) GTFS does not allow multiple trips on the same route at the same time. b) None of these c) Multiple trips are represented as separate routes in GTFS. d) GTFS relies on real-time updates to handle such cases. \\ \hline
    Data Structure& Augmented & What data type is used for the stop\_sequence attribute in GTFS? a) None of these b) Time c) Text d) Integer \\ \hline
 	\bottomrule
    
    \end{tabular}
    \label{table:example_questions}
\end{table*}

To increase task difficulty, we augmented the original 195 MCQs by replacing one answer choice with "none of these," creating 780 additional variants ($195\times 4$). This modification tested the model's ability to handle scenarios where the correct answer might not be explicitly provided. We refer to this experiment as~\texttt{Experiment II}. Table~\ref{table:example_questions} presents ten sample MCQs used in this project.

A potential limitation of~\texttt{Experiment I} is the relatively small number of questions in certain categories, such as Term Definition with only 14 questions. This imbalance could introduce bias into the evaluation. To address this, we expanded the question set to 74 questions per category, resulting in 444 total MCQs ($74 \times 6$) for \texttt{Experiment III}. Similar to \texttt{Experiment II}, we augmented these questions by replacing one answer choice with "none of these," creating 1776 MCQs ($444 \times 4$) for \texttt{Experiment IV}.

\subsection{Experiment for Transportation Information Retrieval}

The transportation information retrieval task assesses an LLM's ability to extract relevant information from a provided dataset. We employed a question-answering (QA) format, differing from the multiple-choice questions used in the Understanding task by omitting potential answer choices. To generate correct responses, the LLM must retrieve information from a given GTFS dataset. The San Antonio VIA GTFS feed, encompassing data for 98 bus routes, served as the foundation for our questionnaire.

Due to LLM context length limitations, we reduced our dataset to three bus routes (Routes 242, 243, and 246), encompassing 34 trips and 60 unique stops. We developed 80 short-answer questions requiring basic search, filtering, sorting, grouping, and joining operations across multiple files. These questions were categorized into simple and complex levels.

\begin{itemize}
    \item \textbf{Simple questions} are based on simple lookups within the same file or two different files (using relational keys) within GTFS, such as \texttt{What \texttt{route\_type} corresponds to \texttt{route\_id 243}?}
 
   \item \textbf{Complex questions} need multiple files to extract information, require a deeper understanding, and could be open-ended. An example may look like \texttt{Tell the \texttt{route\_long\_name} in which there is a \texttt{stop\_name} as "GILLETTE \& PLEASANTON RD."?} To support this question, four data files are needed, namely \texttt{stops.txt}, \texttt{stop\_times.txt}, \texttt{trips.txt}, \texttt{routes.txt}. The LLM needs first to find the stop name \texttt{GILLETTE \& PLEASANTON RD} and the the corresponding \texttt{stopid} from \texttt{stops.txt}. Then, using the \texttt{stopid} to reterive the \texttt{tripId} from \texttt{stop\_times.txt}. Based on the \texttt{tripId}, the LLM can find the \texttt{routeId} from \texttt{trips.txt}. Finally, the \texttt{route\_long\_name} can be found from \texttt{routes.txt} using the \texttt{routeId}.
\end{itemize}
We denote this transportation information retrieval task as \texttt{Experiment V}.

\subsection{Evaluation Methods}
We employ accuracy as the evaluation metric to assess LLM performance across Experiments I to IV. Accuracy is calculated as:
\begin{equation}
    accuracy = \frac{1}{N}\sum_{i=1}^N c_i, 
\end{equation}
where $N$ is the number of questions and $c_i$ is a binary indicator. $c_i=1$, if LLM output $\hat y_i$ for question $i^{th}$ equals the ground true answer, $y_i$ of that question (i.e., $\hat y_i = y_i$); otherwise, $c_i=0$. 

All LLM outputs $\hat y_i$ iare generated using zero-shot learning, where the model responds to questions without prior specific training. For Experiments I-IV, accuracy is determined by exact match between $\hat y_i$ and $y_i$. In Experiment V, accuracy is based on semantic equivalence between $\hat y_i$ and $y_i$.

\section{Experimental Results}
This section presents the evaluation results of the five conducted experiments. The outcomes of Experiments I to IV, focusing on understanding, are detailed in Section~\ref{sec:understanding_result}. The results of Experiment V, which investigates information retrieval, are provided in Section~\ref{sec:understanding_ir}.

\begin{table*}[!tb]
   \normalsize
	\centering
	\caption{LLMs performance on transportation information understanding}
     \resizebox{1\textwidth}{!}{
        \small
        \begin{tabular}{l||p{1.26cm}|p{1.26cm}||p{1.26cm}|p{1.26cm}||p{1.26cm}|p{1.26cm}||p{1.26cm}|p{1.26cm}} \toprule \hline
    	\multirow{2}{*}{\textbf{Categories}} & \multicolumn{2}{c||}{\textbf{Experiment I}} &  \multicolumn{2}{c||}{\textbf{Experiment II}} & \multicolumn{2}{c||}{\textbf{Experiment III}} & \multicolumn{2}{c}{\textbf{Experiment IV}} \\\cline{2-9}

        & \textbf{3.5-turbo} & \textbf{4o} & \textbf{3.5-turbo} & \textbf{4o} & \textbf{3.5-turbo} & \textbf{4o} & \textbf{3.5-turbo} & \textbf{4o} \\\hline\hline
        
        {Term Definition} & \textbf{85.71\%} & \textbf{85.71}\% & 76.69\% & \textbf{80.36\%} & 85.14\% & \textbf{95.95\%} & 68.24\% & \textbf{84.46\%} \\\hline
        
        {File Structure} & \textbf{94.12\%} & \textbf{94.12\%} & 80.88\% & \textbf{88.24\%} & 68.92\% & \textbf{94.95\%} & 57.77\% & \textbf{81.08\%} \\\hline
        
        {Data Structure} & \textbf{90.00\%} & 86.67\% & 74.17\% & \textbf{75.83\%} & 81.08\% & \textbf{89.19\%} & 69.26\% & \textbf{77.03\%} \\\hline
        
        {Common Reasoning} & \textbf{85.71\%} & 78.57\% & 61.61\% & \textbf{75.89\%} & 66.22\% & \textbf{79.73\%} & 52.03\% & \textbf{69.26\%} \\\hline
        
        {Categorial Mapping} & \textbf{54.05\%} & \textbf{54.05\%} & 47.97\% & \textbf{48.65\%} & \textbf{54.05\%} & \textbf{54.05\%} & 47.97\% & \textbf{48.65\%}  \\\hline
        
        {Attribute Mapping} & \textbf{96.88\%} & \textbf{96.88\%} & 82.03\% & \textbf{98.44\%} & 90.54\% & \textbf{91.89\%} & 75.00\% & \textbf{93.58\%}  \\\hline\bottomrule
    	\end{tabular}
     }
	\label{table:result}
\end{table*}

\subsection{Result of Transportation Information Understanding}
\label{sec:understanding_result}

We present the evaluation results of Experiments I-IV, assessing LLM performance on the transportation information understanding task, in this section. Both GPT-3.5-turbo and GPT-4 models were employed, leveraging their respective maximum context lengths of 16,385 and 128k tokens. Table~\ref{table:result} summarizes the overall performance of the four experiments, highlighting the best performance within each category.

\subsubsection{Experiment I vs Experiment III}
Experiments I and III utilized non-augmented questions, with 195 and 444 MCQs, respectively. While GPT-3.5-turbo and GPT-4 demonstrated comparable performance across most categories in Experiment I, GPT-3.5-turbo unexpectedly outperformed GPT-4 in Data Structure and Common Reasoning. This is surprising given GPT-4's established superiority in general natural language processing. However, Experiment III, with its larger dataset, revealed a consistent performance advantage for GPT-4 over GPT-3.5-turbo, with an average improvement of approximately 10\% and a peak improvement of approximately 26\% in the File Structure category. We attribute GPT-4's lower performance in Experiment I to the limited dataset size, particularly evident in the Term Definitions category with only 14 questions. Such a small sample size hinders the reliable assessment of LLM capabilities.

\subsubsection{Augmented Dataset vs Non-Augmented Dataset}
The MCQ sets for Experiments II and IV were generated by augmenting the answer choices of Experiments I and III, respectively, with the option "none of these." This modification increased question difficulty, resulting in a significant decline in LLM performance across most categories compared to their non-augmented counterparts, except for Attribute Mapping. The average performance decrease was approximately 10\%. Notably, GPT-4 demonstrated greater robustness to this increased difficulty, with a performance drop of approximately 6.66\%, compared to GPT-3.5-turbo's decline of 13.23\%, a difference of roughly twofold.

\subsection{Result of Transportation Information Retrieval}
\label{sec:understanding_ir}

This section presents the experimental results of evaluating LLM capabilities in transportation information retrieval (Experiment V). A total of 42 simple and 38 complex short-answer questions were employed. Alongside each question, necessary data files were provided as input. The LLMs were tasked with extracting relevant information from these files to generate accurate responses.

\begin{figure}
    \centering
    \includegraphics[width=0.75\linewidth]{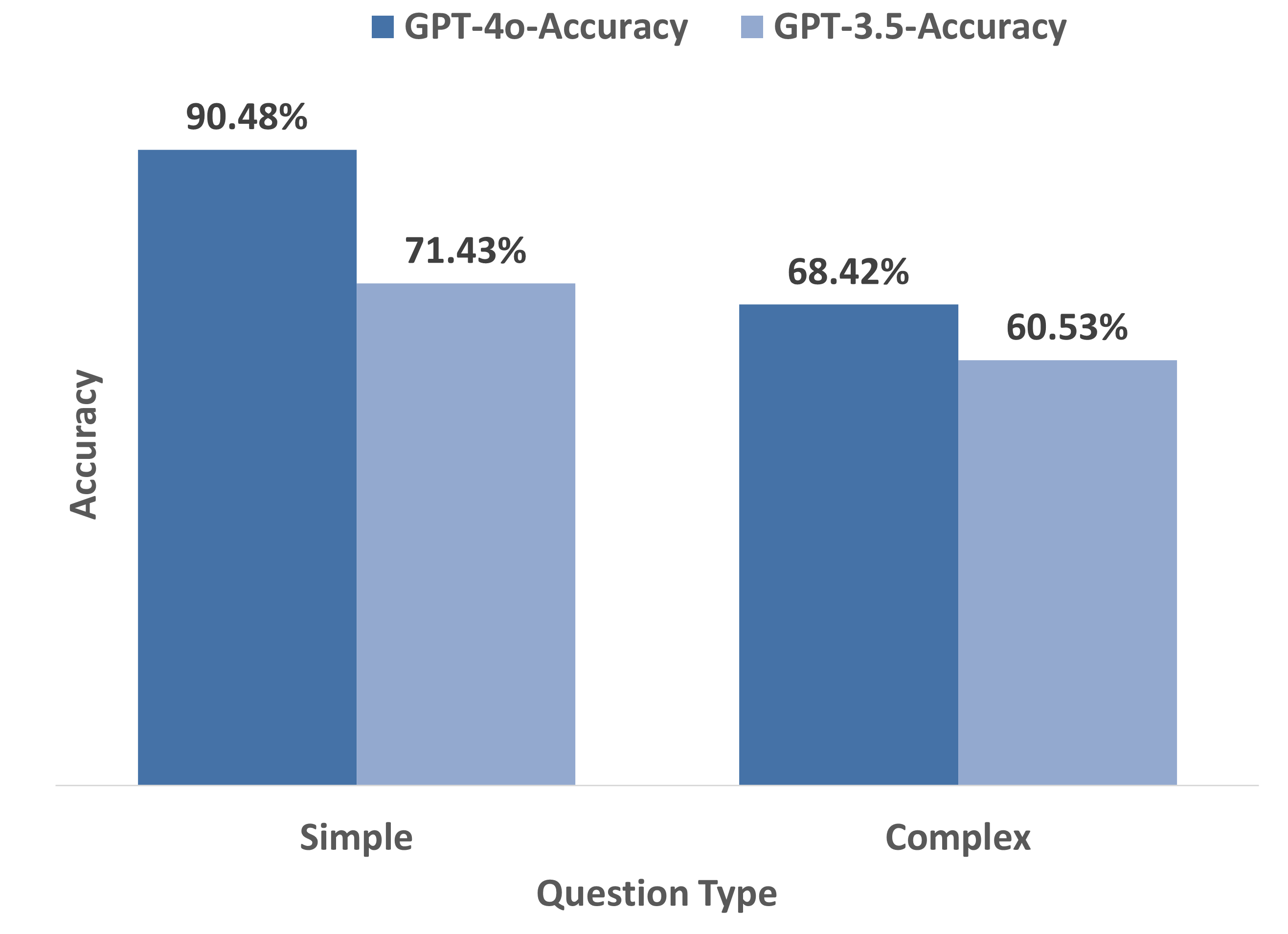}   
    \caption{Summary of performance by question type for GPT 3.5-turbo and GPT-4o on GTFS Retrieval Benchmark.}
    \label{fig:retrieval}
\end{figure}

Figure~\ref{fig:retrieval} demonstrates the significant potential of LLMs for information retrieval tasks. When presented with simple queries, LLMs achieved an impressive average accuracy of approximately 81\%, with a best performance of 90.48\% by GPT-4o. However, performance declined to an average of approximately 64\% when handling complex questions requiring data integration across multiple files.

GPT-4o consistently outperformed GPT-3.5-turbo across all question types, demonstrating an average performance increase of approximately 15\%. This finding aligns with previous results from Experiments II-IV and the comparison between augmented and non-augmented datasets in Section~\ref{sec:understanding_result}, reinforcing GPT-4o's superior capabilities.



 \section{Analysis and Discussion}

\subsection{LLMs for Transportation Applications}
This study explored the potential of large language models in the context of public transportation, focusing on understanding and retrieval tasks within San Antonio's transit system. Our experiments, employing GPT-3.5-turbo and GPT-4, demonstrated varying levels of LLM performance across different tasks on 3275 questions. The model performance is ranging from 47.97\% to 98.44\% accuracy.

A notable challenge emerged in the Categorical Mapping task, where both models achieved relatively low accuracy (an average of 51.35\% and 51.01\%, respectively). This is likely attributed to the semantic similarity between certain categories, as evidenced by the high cosine similarity between ``Rail" and ``Light Rail." Addressing this issue could involve developing more nuanced category embeddings or exploring alternative classification methods.

The augmented understanding tasks (Experiments II and IV) also revealed a significant performance decline, indicating that LLMs struggle with inherent question ambiguity. This suggests a need for more robust question clarification or disambiguation techniques. Additionally, while LLMs excelled at simple information retrieval (Experiment V), their performance deteriorated significantly with complex queries requiring data integration. This highlights the importance of developing strategies to enhance LLM capabilities in handling multifaceted information.

Overall, our findings suggest that LLMs hold promise for transportation applications. However, to realize their full potential, careful engineering and fine-tuning are essential to address the identified challenges. Future research should focus on improving LLM performance in tasks involving semantic similarity, question ambiguity, and complex data integration.

\subsection{Inconsistency Issues in LLM}
Through this work, we also noticed that there may be other potential issues that make the adoption of LLMs in transportation challenging. One of the most notable challenges is performance inconsistency that can manifest in various ways, including contradictions, fluctuating levels of details, and varying degrees of factual accuracy. Here we provide different key aspects of such inconsistency:

\begin{itemize}
    \item Contradictory Responses: LLMs can produce responses that contradict each other when asked the same or similar questions in different contexts or at different times. This is particularly problematic in applications where reliability and coherence are crucial.

\item Context Sensitivity: LLMs sometimes fail to maintain context over long conversations or across multiple interactions. They may provide contextually appropriate answers in one part of a conversation but fail to do so later, leading to inconsistencies.

\item Factual Accuracy: While LLMs can generate text based on a vast amount of data, they may sometimes produce incorrect or misleading information. The same question asked in different ways or at different times can yield different, sometimes contradictory, factual statements.

\item Bias and Fairness: LLMs trained on diverse datasets can inadvertently learn and reproduce biases present in the training data. This can lead to inconsistent and unfair responses that reflect these biases.

\item Detail Level Fluctuation: The detail level in responses can vary, with LLMs sometimes providing overly detailed answers and at other times being too vague. This fluctuation can be problematic for users who rely on consistent levels of information.

\end{itemize}



Additionally, we noticed that OpenAI's Playground and programmatic API may produce inconsistent results, even if both of them use the same pre-trained model. Below are some potential reasons that have been discussed among the OpenAI community\footnote{https://community.openai.com/t/playground-and-api-discrepancies/203672}\footnote{https://community.openai.com/t/is-chatgpt-api-actually-getting-worse/97214/21?page=2}: 

\begin{itemize}
    \item Version and Configuration Differences: The API and Playground might be configured differently in terms of model versions, parameters such as: temperature, maximum tokens, and other settings. Even slight variations in these configurations can lead to different outputs for the same input.

\item Request Formatting: The way requests are formatted and sent can differ between the API and Playground. For instance, users may include different prompts, system messages, or structure their inputs differently, impacting the model's response.

\item Session Management: The Playground often maintains session history, allowing the model to retain context from previous interactions within the same session. In contrast, API requests might be stateless unless explicitly designed to maintain context, leading to potential inconsistencies in responses.

\item User Inputs and Interaction Style: Users might interact differently with the model depending on the interface. The interactive nature of the Playground can lead to more iterative and refined queries, while API usage might involve more straightforward, single-shot requests.

\item Model Updates and A/B Testing: OpenAI may deploy updates or conduct A/B testing on either the API or Playground, leading to temporary differences in model behavior as new features or improvements are tested.

\item Error Handling and Feedback: The Playground often provides more immediate feedback and error handling, helping users adjust their inputs on the fly. The API, being programmatic, might not offer the same level of interactive guidance, which can affect the quality and consistency of responses.
\end{itemize}

We believe understanding these factors is crucial for developers and users to effectively leverage ChatGPT's capabilities and mitigate inconsistency issues. By aligning configurations, maintaining context, and standardizing input formats, users can achieve more consistent results across both the API and Playground interfaces.

\subsection{Prompt Engineering}

Prompt engineering is the systematic design of textual inputs to guide large language models (LLMs) toward desired outputs. By carefully crafting prompts, practitioners can optimize model performance for various tasks, including generation, manipulation, and reasoning. This process is crucial for maximizing the utility of LLMs in diverse applications~\cite{varshney2023an},\cite{aws2024what}. For instance, the performance of our experiments may improve by up to~$\approx8\%$ through optimized prompting, highlighting the necessity of prompt engineering in transportation applications that involve LLMs.

\section{Conclusion and Future Work}
This work evaluates the ability of Large Language Models (LLMs) to understand public transportation information through two tasks: \textit{understanding} and \textit{information retrieval.} The LLMs achieved accuracy ranging from $47.97\%$ to $98.44\%$ on the understanding task and $60.53\%$ to $90.48\%$ on information retrieval. The strong performance on certain understanding tasks, such as Term Definition, File Structure, Data Structure, and Attribute Mapping, indicates that pre-trained LLM models have gathered a substantial amount of transportation-related information from their training datasets. However, the considerable disparity between the best and worst-performing tasks suggests that these models may have been trained on an imbalanced dataset, with significantly less information available on specific topics. While they can handle tasks involving unknown data—evidenced by their high performance in information retrieval—their effectiveness appears to diminish as task complexity increases. 
This study highlights the significant potential of large language models in transforming public transit systems and enhancing user experiences. By improving passenger information services, operational efficiency, and accessibility, LLMs present a variety of applications that can greatly benefit public transit. However, the notable performance gaps between the best and worst tasks must be addressed before real-world implementation. Additionally, it will be crucial to tackle ethical concerns and ensure responsible use of these technologies as the field evolves. With ongoing research and development, LLMs could play a vital role in the future of public transportation.

\section*{Acknowledgment}
 This work is supported by the National Science Foundation under Grant No. 2131193. Any opinions, findings, conclusions, or recommendations expressed in this material are those of the author(s) and do not necessarily reflect the views of the National Science Foundation.

\bibliographystyle{IEEEtran}  
\bibliography{bibfile}

\end{document}